%% file: main.tex
\documentclass[10pt,twocolumn,letterpaper]{article}

\usepackage{iccv}              
\usepackage{times}
\usepackage{epsfig}
\usepackage{graphicx}
\usepackage{amsmath}
\usepackage{amssymb}
\usepackage{wrapfig}
\usepackage{multirow}
\usepackage{makecell}
\usepackage{caption}
\usepackage{pifont}
\usepackage{xcolor}
\definecolor{darkgreen}{rgb}{0.0,0.6,0.0}
\usepackage[symbol]{footmisc}
\usepackage[linesnumbered,ruled]{algorithm2e}
\usepackage{float}

\input{preamble}

\definecolor{iccvblue}{rgb}{0.21,0.49,0.74}
\usepackage[pagebackref,breaklinks,colorlinks,allcolors=iccvblue]{hyperref}

\def\paperID{4573} 
\def\confName{ICCV}
\def\confYear{2025}

\title{LookOut: Real-World Humanoid Egocentric Navigation}

\author{Boxiao Pan \qquad Adam W. Harley \qquad C. Karen Liu$^{\footnote[1]{}}$ \qquad Leonidas J. Guibas$^{\footnote[1]{}}$\\
\\
\text{Stanford University}
}
\begin{document}

\maketitle

\footnotetext[0]{* Equal advising}

\input{content/main_paper/text/0_abstract}
\input{content/main_paper/text/1_intro}
\input{content/main_paper/text/2_related_works}
\input{content/main_paper/text/3_method}

\input{content/main_paper/text/4_data}
\input{content/main_paper/text/5_experiments}
\input{content/main_paper/text/6_conclusion}
\input{content/main_paper/text/7_acknowledgment}

\clearpage
{
    \small
    \bibliographystyle{ieeenat_fullname}
    \bibliography{main}
}

\end{document}

%% file: preamble.tex
\usepackage{booktabs}
\usepackage{tabularx}
\usepackage{wrapfig}
\usepackage{enumitem}
\usepackage{xspace}

\newcommand{\parahead}[1]{\noindent\textbf{#1}.\ }
\renewcommand{\paragraph}[1]{{\vspace{1mm}\noindent \bf #1}.}

\newcommand{\name}{LookOut\xspace}

%% file: content/main_paper/text/0_abstract.tex
\begin{abstract}
The ability to predict collision-free future trajectories from egocentric observations is crucial in applications such as humanoid robotics, VR / AR, and assistive navigation. In this work, we introduce the challenging problem of predicting a sequence of future 6D head poses from an egocentric video. In particular, we predict both head translations and rotations to learn the active information-gathering behavior expressed through head-turning events. To solve this task, we propose a framework that reasons over temporally aggregated 3D latent features, which models the geometric and semantic constraints for both the static and dynamic parts of the environment. Motivated by the lack of training data in this space, we further contribute a data collection pipeline using the Project Aria glasses, and present a dataset collected through this approach. Our dataset, dubbed Aria Navigation Dataset (AND), consists of 4 hours of recording of users navigating in real-world scenarios. It includes diverse situations and navigation behaviors, providing a valuable resource for learning real-world egocentric navigation policies. Extensive experiments show that our model learns human-like navigation behaviors such as waiting / slowing down, rerouting, and looking around for traffic while generalizing to unseen environments. Check out our project webpage at \url{https://sites.google.com/stanford.edu/lookout}.
\vspace{-3mm}
\end{abstract}

%% file: content/main_paper/text/1_intro.tex
\section{Introduction}

\begin{figure*}[t]
    \centering
    \includegraphics[width=\linewidth]{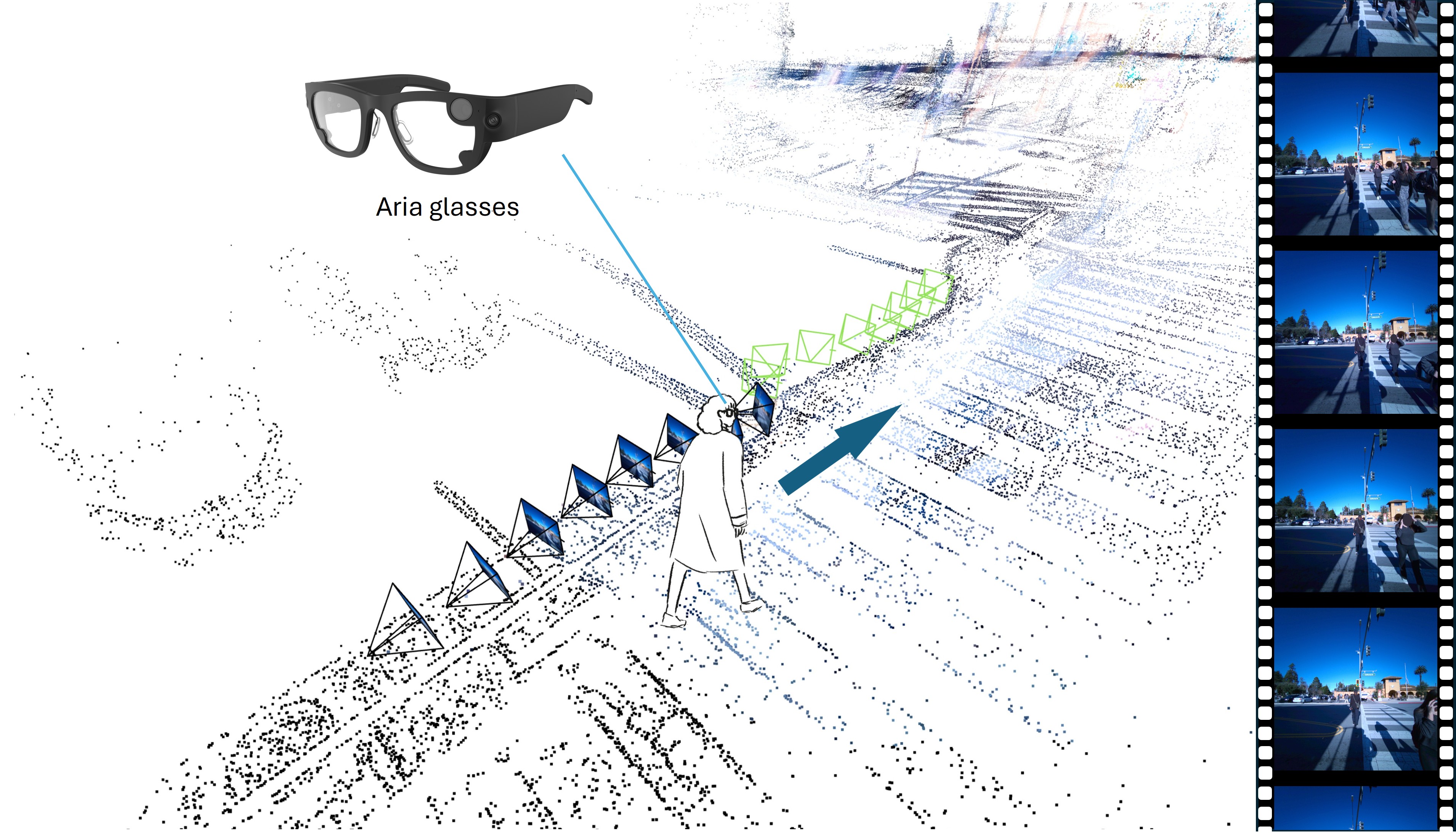}
    \caption{
    \textbf{Problem formulation}. Given a posed egocentric video (black-outlined frustums, with frames shown in detail on the right), our model predicts a sequence of 6D head poses in the future (green-outlined frustums). We design a data collection pipeline with the Project Aria glasses and train our model on a dataset collected this way. This problem features real-world navigation challenges including collision avoidance with static and dynamic obstacles, and human-like information-gathering behaviors (\eg looking to the sides when crossing roads in this example). The point cloud is shown for visualization but is not an input to the model.
    }
    \vspace{-3mm}
    \label{fig:teaser}
\end{figure*}

Navigating safely in the real world from egocentric observations is an ability that we humans possess, yet extremely difficult for machines to learn. This is largely due to the diverse and complex situations that exist in practical scenarios. Such capabilities are crucial for various applications including humanoid robotics~\cite{HumanoidNav2024}, VR / AR~\cite{EgoCast2024}, and assistive navigation~\cite{EgoNav2024}.  

Many works have approached this problem from different angles. Vision-Language Navigation (VLN)~\cite{HumanAwareVLN2025,GoatBench2024,OVON2024,ObjNav12023,LangNav2023} focuses on localizing goals and planning long-term goal-directed paths, typically in simulated static environments. Robotic social navigation~\cite{VLMSocialNav2024,SocialLLaVA2024,HumanLikeNav2023,Sacson2023,SCAND2022} learns socially compliant navigation policies in dynamic environments. These works generally target wheeled or legged navigation robots, whose action and observation distributions are vastly different from the human form factor. Recently, several works investigate egocentric navigation that predicts trajectories~\cite{EgoNav2024} or full-body poses~\cite{EgoCast2024} for humans. They however assume static environments.

Despite these advances, a real-world deployable humanoid egocentric navigation policy remains challenging. First, a method for humanoid navigation in dynamic environments is lacking. Second, existing methods ignore an important aspect of human-like navigation, which is the active information gathering via head turning. Humans often rotate their heads and look for useful information. For example, we look to the side before crossing roads to check any passing vehicles, look downward when stepping off / onto curbs, etc. This ability is helpful for real-world deployment, partly due to the limited FoV of cameras. Lastly, we do not have a way to collect multi-modal labeled training data at scale due to the difficulty of deploying humanoid robots in the real world.

In this work, we make steps towards a real-world deployable humanoid navigation policy from all three of these fronts. To tackle the challenge of 3D dynamic scene awareness, we propose a model that unprojects per-frame DINO~\cite{DINO2021,DINOv22023} features to 3D, and aggregates the 3D feature volumes across time to gain a holistic understanding of the geometric constraints posed by the environment. Moreover, through training on our collected dataset that contains extensive dynamic obstacles (\ie pedestrians and vehicles), our model effectively learns the ability to navigate around both static and dynamic objects. To model active information gathering, we design our framework to predict 3D head rotations in addition to translations (\ie 6D head pose prediction), which can be used to calculate velocity commands normally input to humanoid robots~\cite{NextToken2024,Navila2024,ChallengingTerrain2024}. Additionally, in our data collection process, we ask our human subjects to follow a careful information-gathering strategy, \eg always looking for passing cars before crossing roads. To solve the challenge of collecting useful data at scale, we propose a data collection pipeline that uses a pair of Project Aria glasses~\cite{Aria2023} as the data collection tool. This pipeline enables naturalistic human navigation demonstration collection without drawing attention~\cite{HumanoidNav2024}. Unlike traditional collection pipelines that require carefully mounting various sensors~\cite{HumanoidNav2024} or teleoperating robots~\cite{SCAND2022}, our pipeline is extremely easy to set up, taking only a few seconds at the beginning of each recording session, while providing various data modalities including RGB videos, audio, eye gaze, SLAM reconstructed head poses and point cloud. It thus provides a way to scale up the data collection process with minimal effort. With this pipeline, we collected a dataset that consists of 4 hours of real-world navigation sessions, covering 18 densely populated places.

In summary, we make the following contributions: (1) we introduce the challenging task of 6D head pose trajectory prediction from posed egocentric observations, under the presence of static and dynamic obstacles. (2) We propose a model dubbed \name that aggregates unprojected 3D DINO features over time for semantic and geometric understanding, which proves effective in solving the task. (3) We contribute a data collection pipeline that leverages a pair of Project Aria glasses and requires minimal effort, providing a way to scale up the data collection effort at ease. (4) We collected a dataset with this pipeline, which consists of 4 hours of real-world navigation sessions and covers 18 places with dense and diverse traffic.

%% file: content/main_paper/text/2_related_works.tex
\section{Related Work}
\label{sec:related_work}

\parahead{Vision-language navigation}
Studies on the task of Vision-Language Navigation (VLN) are arguably the most prevalent in the vision community around embodied navigation. This task is defined roughly as navigating to a specified goal location. Depending on the goal specification, the task has several variants as Point-Goal Navigation (PointNav)~\cite{PointNav2021}, Object-Goal Navigation (ObjectNav)~\cite{ObjNav12023,ObjNav22023,OVON2024}, Image-Goal Navigation (ImageNav)~\cite{ImageNav2023}, Language-Goal Navigation (LangNav)~\cite{LangNav2020,LangNav2023}, and Audio-Visual Navigation~\cite{SoundSpaces2020}. Some works also propose frameworks that unify several specifications~\cite{GoatBench2024,HumanAwareVLN2025}. These works are generally developed in simulated environments, and focus on long-term path planning where no or simple~\cite{HumanAwareVLN2025} dynamic obstacles exist. In this work, we focus on short-term navigation whose main objective is collision-free locomotion. 

\parahead{Robotic social navigation}
Classical methods have extensively studied goal-conditioned path planning for robots~\cite{RobotPathPlan2021} that develop mathematical solutions given almost perfect knowledge on the environment. More relevant to our work are works that study egocentric navigation for robots. Robotic social navigation aims to predict a collision-free path~\cite{VLMSocialNav2024,LifeLongNav2024,LiDARLegged2024} or higher-level instructions~\cite{TimeToCollision2019,SocialLLaVA2024} for robots, given egocentric sensing input such as RGB, LiDAR, and odometry. Common datasets for this task are collected by teleoperating robots~\cite{SCAND2022} or human collectors wearing a sensor suite~\cite{HumanLikeNav2023}. These works generally target wheeled or legged navigation robots. The actions and observations of such robots are drastically different from those of humanoids, due to their smaller scale, faster speed, and different morphologies. On the other hand, studies for humanoid robots~\cite{LaserHumanoid2012,EffiHumanoidLoco2018,StereoHumanoid2004,HumanoidObstacle2011,HumanoidVisionObs2013} generally design classical methods with laser and stereo vision input for simplified environments. 

\parahead{Egocentric navigation for humans}
Significant progress has been made in egocentric human motion estimation~\cite{HMD22024,EgoSense2024,EgoEgo2023,PDEgo2019,EgoCap2016,XREgoPose2019,EgoPoser2024,You2Me2020}, while forecasting future motion or trajectories that takes environment constraints into account is much less explored. COPILOT~\cite{COPILOT2023} predicts human-environment collision from multi-view egocentric videos in the form of collision labels and heatmaps. EgoNav~\cite{EgoNav2024} uses a diffusion model to forecast future trajectories from a chest-mounted RGBD camera input and past trajectories. EgoCast~\cite{EgoCast2024} predicts future full-body poses from a head-mounted RGB camera input and past head trajectories. Notably, EgoNav focuses on navigation while EgoCast studies diverse social and skilled human activities. However, all three works assume static environments, and do not learn the active information-gathering behavior critical in real-world scenarios. Moreover, we contribute a data pipeline that can be easily deployed at scale.

\parahead{Egocentric datasets for humans}
To study this task, we need a dataset that records real-world human navigation scenarios with significant presence of both static and dynamic obstacles, and provides data modalities on egocentric RGB videos, 6D head poses, and preferably also scene point cloud for collision checking. Traditional egocentric video datasets~\cite{EPICKITCHENS2018,Ego4D2021,EgoExo2021,CharadesEgo2018} are generally captured in the monocular setting, and thus do not have camera pose annotations. The activities in these datasets are also diverse and do not focus on navigation. Some works~\cite{COPILOT2023,EgoGen2024} propose synthetic data generation pipelines to simulate virtual humans walking in synthetic scenes. However, the generated motions are simple and unnatural due to the limitation of human motion generation methods. Autonomous driving dataset such as Waymo Open~\cite{WaymoOpen2020} comprehensively covers real-world traffic scenarios and provide dense 3D tracking for pedestrians and vehicles, but they do not record egocentric data for pedestrians. Recently, the Project Aria glasses~\cite{Aria2023} emerged as a convenient and natural way to record egocentric data. In particular, the Aria Machine Perception Service (MPS) provides an easy and highly optimized method to obtain accurate 6D camera (head) pose trajectories, environment point clouds, eye gazes, and more. Meta released several datasets collected with the Project Aria, and among them Aria Everyday Activities (AEA)~\cite{AEA2024} and Nymeria~\cite{Nymeria2024} record diverse indoor and outdoor activities. However, they either feature only single-human activities~\cite{AEA2024} or have multiple actors only in collaborative activities~\cite{Nymeria2024}, so real-world navigation scenarios with potentially colliding agents are not captured.

%% file: content/main_paper/text/3_method.tex
\section{Method}
\label{sec:method}

\begin{figure*}[t]
    \centering
    \includegraphics[width=\linewidth]{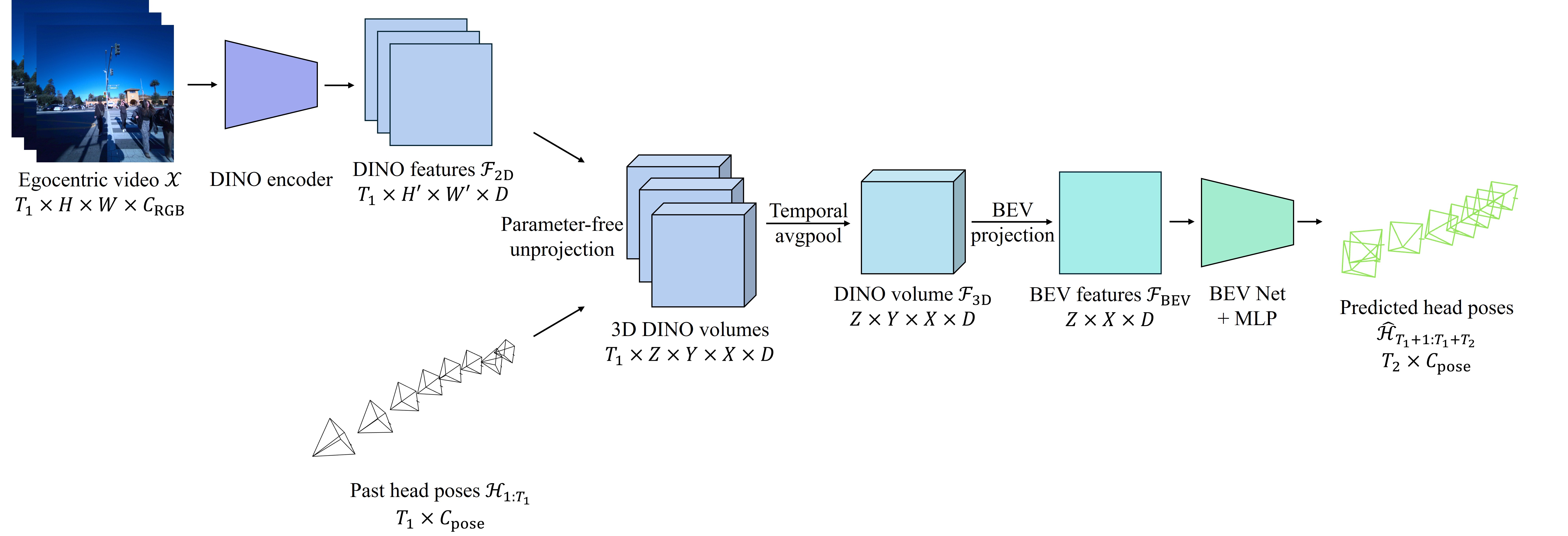}
    \caption{
    \textbf{\name architecture}. Given a posed egocentric video, we obtain frame-wise DINO features with the pre-trained encoder, and unproject them to 3D for temporal aggregation. The aggregated features are then projected to BEV for further processing and eventually used to predict future head poses.
    }
    \vspace{-3mm}
    \label{fig:framework}
\end{figure*}

\subsection{Problem Formulation}
\label{subsec:prob_def}
We illustrate our problem formulation in~\cref{fig:teaser}. Given a posed egocentric video $\mathcal{X} \in \mathbb{R}^{T_1 \times H \times W\times 3}$ and $\mathcal{H}_{1:T_1} =\{\mathbf{h}_1, \mathbf{h}_2, \cdots, \mathbf{h}_{T_1}\} \in \mathbb{R}^{T_1 \times 9}$, our goal is to predict the 6D head pose sequence for a short period in the future, \ie $\mathcal{H}_{T_1+1:T_1+T_2} =\{\mathbf{\hat{h}}_{T_1+1}, \mathbf{\hat{h}}_{T_1+2}, \cdots, \mathbf{\hat{h}}_{T_1+T_2}\} \in \mathbb{R}^{T_2 \times 9}$. The head poses, which are also the camera poses, are parameterized as $\mathbf{h}_t = [\mathbf{t}_t | \mathbf{r}_t]$, where the rotation component $\mathbf{r}_t$ adopts the 6D continuous rotation representation~\cite{6D2019}. Throughout our experiments, $T_1 = T_2 = 8$. The head poses are defined in a head-centered canonical frame specified in~\cref{subsec:implementation}. 

\subsection{Model}
\label{subsec:model}
The core functionality our model needs to have is extracting semantic and geometric information for the surrounding environment from a single monocular video stream. This prompts us to consider the following questions: (1) a strong visual encoder for semantic modeling, and (2) a way to aggregate information and reason in the 3D space. These motivate our key design choices of the model. For (1), we use the pre-trained DINO~\cite{DINO2021,DINOv22023} encoder to extract per-frame feature maps, due to its strong open-vocabulary semantic encoding capabilities~\cite{DINOTracker2024,JacobiNerf2023,DecomposeNerf2022,NFFF2022}. For (2), we adopt a strategy used in a number of object and scene representation models named ``Parameter-Free Unprojection"~\cite{SimpleBEV2023,GeometryRNN2018,DeepVoxels2019,Predictive3DMap2019,CommonSense2019}. Specifically, we bilinearly sample a subpixel 2D DINO features for each 3D coordinate defined in the canonical frame to obtain a 3D DINO feature volume. We then temporally aggregate the volumes for all time steps. These design choices endow our model with strong semantic and geometric reasoning capability while not having to rely on explicit geometric sensing input, such as depth and LiDAR. We illustrate our model in~\cref{fig:framework} and describe each component in detail next.

\parahead{DINO feature encoding}
We use the pre-trained DINO2~\cite{DINOv22023} variant \texttt{dinov2\_vits14\_reg}~\footnote{https://github.com/facebookresearch/dinov2}, and apply it to each input frame (down-sampled to $224 \times 224$ spatial resolution). This gives us a temporal sequence of 2D DINO features $\mathcal{F}_{\text{2D}} \in \mathbb{R}^{T_1 \times 16 \times 16 \times 384}$.

\parahead{Parameter-free unprojection}
Following~\cite{SimpleBEV2023,Predictive3DMap2019}, we first define a voxel grid of 3D points in the canonical frame, and project these points to each input frame's pixel space. The feature encoding for each point is then obtained by bilinearly interpolating the 2D DINO feature map. This yields a sequence of 3D DINO feature volumes, which are subsequently aggregated across time for a single 3D feature volume $\mathcal{F}_{\text{3D}} \in \mathbb{R}^{Z \times Y \times X \times 384}$, where $Z=X=96, Y=32$ are the spatial resolutions of the voxel in our Y-up canonical frame. We simply use average pooling across time as the temporal aggregation method.

\parahead{BEV projection}
Directly reasoning in the 3D feature space is expensive and in many cases sub-optimal (as ablated later). We hence project the 3D feature volume obtained above to the ``Bird's Eye View" (BEV) by ``squeezing" the up-axis (Y axis), following~\cite{SimpleBEV2023,Predictive3DMap2019}. The squeezing is done with an MLP that projects the flattened up-channel dimension ($384 \times Y$) to the same-sized channel dimension ($384$). After this step, we end up with a BEV feature map $\mathcal{F}_{\text{BEV}} \in \mathbb{R}^{Z \times X \times 384}$.

\parahead{BEV Net}
$\mathcal{F}_{\text{BEV}}$ is the condensed feature embedding on which we perform the bulk of the computation. Following previous designs~\cite{Predictive3DMap2019,SimpleBEV2023,LiftSplatShoot2020}, our BEV Net consists of 11 sequential \texttt{BEV\_module}s, where each module applies a 2D convolution, a LayerNorm, and an MLP with a GELU activation. The hidden dimension starts at 384 and doubles twice in the middle, reaching the final feature dimension of 1540 while reducing the spatial dimension to $3\times3$.

\parahead{Trajectory prediction}
We first perform a spatial average pooling on the features from the BEV Net, and then use a 3-layer MLP with LayerNorm and GELU activation to get the predicted future head pose sequence $\hat{\mathcal{H}}_{T_1+1:T_1+T_2}$. 

\parahead{Loss function}
We supervise the model with combined L1 losses on the translations and rotations, following~\cite{EgoEgo2023}:
\begin{equation}
    \mathcal{L} = \frac{1}{T_2} \cdot \sum_{t=T_1+1}^{T_1 + T_2} \lambda_{\text{trans}} \cdot ||\mathbf{t}_t - \mathbf{\hat{t}}_t||_1 + \lambda_{\text{rot}} \cdot ||\mathbf{R}_t \mathbf{\hat{R}}_t - \mathbf{I}||_1 
\end{equation}
where $R$ is the rotation matrix converted from the 6D rotation representation, and $I$ is the identity matrix. In our experiments we use $\lambda_{\text{trans}} = \lambda_{\text{rot}} = 1$.

\subsection{Implementation Details}
\label{subsec:implementation}

\parahead{Head-centered canonical frame}
Because we do not input the past head poses to the model (they are only used in the unprojection process), we need to define a canonical frame relative to the current head pose $\mathbf{h}_{T_1}$. Such canonical frames have been widely adopted in prior works~\cite{HuMoR2021,EgoEgo2023,EgoSense2024,ContinualMotion2024,HMD22024}. Following~\cite{ContinualMotion2024}, we also define our frame to be parallel to the ground plane and facing forward, but centered on the head instead. This lets the model operate in a space facing the current heading direction and predict future head poses in a relative sense.  

\parahead{Training details}
We use the AdamW~\cite{AdamW2017} optimizer, and apply a weight decay of 0.05 to all model parameters except biases. We train our model for 700k steps, and use the OneCycle learning rate scheduler~\cite{OneCycleLR2019} with the linear annealing strategy and \texttt{pct\_start} set to 0.05. We use a batch size of 4. The training concludes in about 4 days on a single NVIDIA RTX A6000 GPU.

%% file: content/main_paper/text/4_data.tex
\section{Aria Navigation Dataset (AND)}
\label{sec:dataset}

As discussed in~\cref{sec:related_work}, there are no suitable datasets for our task to the best of our knowledge. We hence design a data collection pipeline to collect our own dataset, which we name as the Aria Navigation Dataset (AND). We next present key data collection steps and dataset statistics.

\subsection{Data Collection Pipeline}
\label{subsec:data_collection}

\parahead{Hardware}
Our data collection hardware consists of only a pair of the Project Aria glasses~\cite{Aria2023}, which has several key advantages of being lightweight, non-intrusive, cheap, and easy-to-set-up compared to prior works that deploy self-built sensor suites~\cite{HumanLikeNav2023,EgoNav2024} or teleoperate robots~\cite{SCAND2022}.

\parahead{Recording process}
Project Aria comes with a mobile app named Aria Studio that allows easy data recording by interacting with the mobile app once prior to each recording session. The app provides selections on recorded data modalities, and in our pipeline, we activate the RGB, SLAM (two monochrome cameras), and eye-tracking cameras, as well as IMU sensor, barometer, and GPS. The cameras all operate at 20fps. Before each recording session, the human subject selects this saved recording profile and starts recording while walking around, without pre-defined instructions or scripts. To capture consistent information-gathering behavior, we instruct the subjects to follow a careful navigation behavior, \eg always checking for passing vehicles before crossing roads. 

\parahead{Data processing}
The raw recorded data comes in a compressed format called VRS~\footnote{https://facebookresearch.github.io/vrs/docs/Overview/}. We run the Aria Machine Perception Services~\footnote{https://facebookresearch.github.io/projectaria\_tools/docs/ARK/mps} to get processed data modalities including 6D head pose trajectories and scene point clouds. The raw RGB frames are distorted due to the fisheye camera, so we undistort them to use as input to our model. The original sequences are further segmented into $(T_1+T_2)$-long clips in a sliding window fashion, with a stride and dilation factor of 6 frames. At 20fps, each clip covers $(8 + 8 - 1) \times 6 / 20 = 4.5$ seconds. During SLAM, points on the dynamic objects are filtered. We apply another filtering process on the reconstructed point cloud to remove noisy points.

\parahead{Privacy}
We have taken measures to follow Project Aria research guidelines. We also use the SOTA de-identification algorithm~\cite{EgoBlur2023} to blur faces in all videos.

\subsection{Dataset Statistics}

\parahead{Locations}
Since we want to capture real-world navigation scenarios where humans need to avoid collision with both static and dynamic obstacles, we selected diverse locations with dense traffic both indoors and outdoors. We picked 18 densely populated places from university campuses, city downtowns, parks, and so on. Many of these locations are expansive, providing great diversity on captured data. We specifically chose times when dense traffic usually happens for data recording, \eg after classes. We also diversified the time-of-the-day distribution.

\parahead{Data scale}
We recorded about 4 hours of data, resulting in 274k RGB frames and 36k clips after processing. 

%% file: content/main_paper/text/5_experiments.tex
\section{Experimental Results}

\begin{table*}[t!]
    \centering
    {%
    \setlength{\tabcolsep}{4pt}
    \begin{tabular}{l|cccccccc}
        \toprule
        \textbf{Method} & L$_1$\_trans $\downarrow$ & L$_1$\_rot $\downarrow$ & Col\_stt\_15 $\uparrow$ & Col\_dyn\_15 $\uparrow$ & Col\_stt\_25 $\uparrow$ & Col\_dyn\_25 $\uparrow$ & Col\_stt\_35 $\uparrow$ & Col\_dyn\_35 $\uparrow$ \\
        \midrule
        Const\_Vel & 0.41 & 0.77 & 85.5 & 91.2 & 80.0 & 81.3 & 74.1 & 73.1 \\
        Lin\_Ext & 0.45 & 1.21 & 86.5 & 92.3 & 77.6 & 82.1 & 73.3 & 72.9 \\
        EgoCast~\cite{EgoCast2024} & 0.34 & 0.63 & 90.5 & 94.6 & 84.6 & 86.2 & 77.4 & 77.8 \\
        Ours & \textbf{0.17} & \textbf{0.16} & \textbf{91.3} & \textbf{97.2} & \textbf{85.6} & \textbf{90.3} & \textbf{79.9} & \textbf{83.1} \\
        GT & 0 & 0 & 92.7 & 97.7 & 88.9 & 93.0 & 83.6 & 85.1 \\
        \midrule
        A$^{*}$+Lin\_Ext & 0.24 & 1.21 & \textbf{98.8} & 82.4 & \textbf{100.0} & 76.5 & \textbf{100.0} & 61.9 \\
        Ours (+goal) & \textbf{0.11} & \textbf{0.15} & 91.7 & \textbf{97.2} & 86.3 & \textbf{91.4} & 82.0 & \textbf{84.6} \\
        \bottomrule
    \end{tabular}
    }
    \caption{\textbf{Comparison with baselines.} Our model outperforms comparable methods on both trajectory prediction and collision avoidance.
    }
    \label{table:baselines}
    \vspace{-3mm}
\end{table*}

We compare \name to baselines quantitatively in~\cref{subsubsec:baselines}. In~\cref{subsubsec:ablation}, we ablate our key design choices. We then present qualitative evaluation results in~\cref{subsec:qual_eval}, which showcase the diverse behaviors our model learns in real-world navigation scenarios. Finally, we investigate failure cases from our model and discuss limitations in~\cref{subsec:failure_cases}. All results are obtained on a held-out set with environments unseen during training. We encourage readers to check our project webpage, which contains video versions of~\cref{fig:qualitative} obtained by continuously rolling out our model given each incoming frame, similar to how it would operate in practice. 

\subsection{Quantitative Evaluation}
\label{subsec:quant_eval}

\parahead{Metrics}
We first evaluate head pose prediction accuracy through the same error function we used for training, \ie the L1 losses on translation (\texttt{L$_1$\_trans}) and rotation (\texttt{L$_1$\_rot}). In order to measure collision with the environment, we define a non-collision score for the static (\texttt{Col\_stt\_k}) and dynamic (\texttt{Col\_dyn\_k}) obstacles, respectively. The score measures the percentage of predictions that are at least $k$ centimeters away from the closest obstacle. For static obstacles, we measure the closest distance from the predicted head translation to the SLAM reconstructed point cloud. While for dynamic obstacles, we first use a monocular metric depth estimation method Depth Pro~\cite{DepthPro2024} to estimate a depth map for each frame in our dataset, and subsequently use DINOv2 + Mask2Former segmentation head~\cite{DINOv22023} to get per-frame semantic segmentation masks. We then take the minimum estimated metric depth values among all pixels labeled ``person" as the closest distance. 
\texttt{Col\_$^{*}$\_avg} is the average value over all $k\in\{15, 25, 35\}$. 
Note that the non-collision score is a rough proxy for collision avoidance, and we also report the values for ground-truth sequences (\texttt{GT}) for reference.

\subsubsection{Comparison with Baselines}
\label{subsubsec:baselines}

\parahead{Baselines}
Since we study a novel problem, there is no directly comparable prior work. As mentioned, the closest prior works to ours are EgoNav~\cite{EgoNav2024} and EgoCast~\cite{EgoCast2024}. EgoNav is an Arxiv preprint that did not release code. We hence adapt EgoCast to our setting. The core part of EgoCast is a transformer-based forecasting module that predicts future 3D full-body poses given past full-body poses and optionally the past egocentric video. To deal with the issue that full-body poses are often not available in practice, it further implements an estimation module to estimate the current frame's full-body pose from past 6D head poses and egocentric video. Both stages are supervised with 3D full-body poses, which our dataset does not have. We hence repurpose the forecasting module to take past head poses (instead of full-body poses) and the egocentric video, and predict future head poses too. We then remove the estimation module. We train it on the same training split as our model.

We additionally implement the following baselines that operate on past head poses: (1) Constant Velocity (\texttt{Const\_Vel}) that uses the linear and angular velocity calculated from the last two input steps to extrapolate future head translations and rotations, (2) Linear Extrapolation (\texttt{Lin\_Ext)} that fits a linear regression model for the past translation and rotation sequences and predicts into the future, and (3) A$^{*}$+Linear Extrapolation (\texttt{A$^{*}$+Lin\_Ext}) that uses linear extrapolation for rotations, but implements an A$^{*}$ algorithm for translations. Specifically, we discretize the space by turning the SLAM reconstructed point cloud into an occupancy grid, using the same spatial resolution as our model. We then take the ground-truth head translation from the last step in the future $T_1+T_2$ as the goal. We also define a maximum velocity that roughly matches the human movement capacity. We use a variant of our model that also takes such goal position as input (directly concatenated to the final MLP) for a fair comparison with this baseline.

\parahead{Analyses}
The comparison results are reported in~\cref{table:baselines}. In the no-goal setting (top), our model achieves the best performance across all metrics, predicting accurate head poses while avoiding collision with both static and dynamic obstacles reliably. When provided with the goal, \texttt{A$^{*}$+Lin\_Ext} achieves near-perfect non-collision scores for static obstacles, because they are explicitly modeled with the scene occupancy (where each voxel grid represents about 600cm$^3$ of space) and the search algorithm basically guarantees a path around occupied regions. However, this baseline does poorly in avoiding dynamic obstacles since they are not represented in the point cloud.

\subsubsection{Ablation Study}
\label{subsubsec:ablation}

We ablate input data modalities and key model designs and summarize the results in~\cref{table:ablation}. 

\parahead{Multi-modal support}
Our model can be easily extended to incorporate additional sensor modalities, \eg depth and point cloud. Specifically, we first convert depth to a point cloud, and then turn it into an occupancy voxel and concatenate with the 3D DINO volume $\mathcal{F}_{\text{3D}}$.
As expected, incorporating these modalities directly relevant to obstacle proximity improves non-collision metrics, which aligns with the findings in previous works~\cite{COPILOT2023}. Using depth especially helps with avoiding dynamic obstacles because the SLAM reconstructed point cloud only contains static objects.

\parahead{Model design}
We first validate the effectiveness of the DINO feature encoding by ablating a variant that unprojects raw RGB frames without passing through DINO (\texttt{w/o DINO}). As shown, DINO features contribute significantly to our model's performance due to its strong semantic feature encoding capabilities. We next inspect the impact of having the intermediate 3D feature space, for which we ablate a variant that temporally pools over 2D DINO features $\mathcal{F_{\text{2D}}}$ instead (\texttt{2D Only}). We can see that the 3D feature space improves performance by granting an explicit geometric notion to the features. Finally, we investigate whether the BEV projection is beneficial by comparing against directly applying 3D convolutions on $\mathcal{F}_{\text{3D}}$ (\texttt{3D Conv}). This variant performs on par to our model, but is more computationally expensive due to 3D convolutions. 

\begin{table}[t]
    \centering
    \resizebox{\linewidth}{!}
    {%
    \setlength{\tabcolsep}{4pt}
    \begin{tabular}{l|cccc}
        \toprule
        \textbf{Method} & L$_1$\_trans $\downarrow$ & L$_1$\_rot $\downarrow$ & Col\_stt\_avg $\uparrow$ & Col\_dyn\_avg $\uparrow$ \\
        \midrule
        PCD only & 0.40 & 0.88 & 83.2 & 84.6 \\
        RGB+PCD & 0.17 & 0.14 & \textbf{87.8} & 90.1 \\
        Depth only & 0.22 & 0.23 & 87.0 & \textbf{91.6}  \\
        RGB+Depth & \textbf{0.15} & \textbf{0.13} & 87.4 & 91.4 \\
        \midrule
        w/o DINO & 0.35 & 0.67 & 84.5 & 85.3 \\
        2D Only & 0.26 & 0.44 & 84.9 & 86.2 \\
        3D Conv & \textbf{0.17} & 0.19 & \textbf{85.6} & 89.9 \\
        Ours & \textbf{0.17} & \textbf{0.16} & \textbf{85.6} & \textbf{90.2} \\
        GT & 0 & 0 & 88.4 & 91.9 \\
        \bottomrule
    \end{tabular}
    }
    \caption{\textbf{Ablation study.} Each design choice helps performance. 
    }
    \label{table:ablation}
    \vspace{-3mm}
\end{table}

\subsection{Qualitative Evaluation}
\label{subsec:qual_eval}

\begin{figure*}[t]
    \centering
    \includegraphics[width=\linewidth]{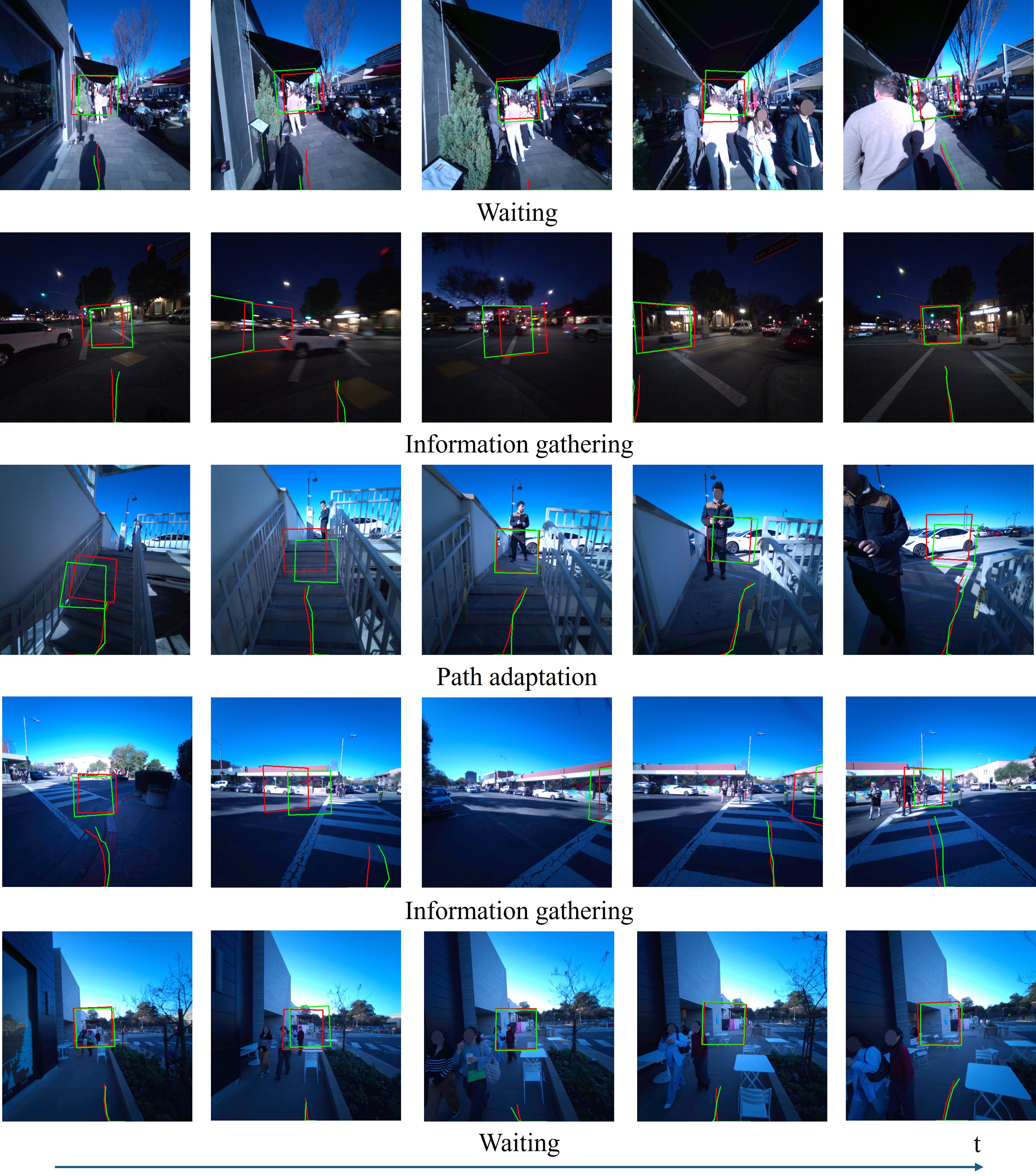}
    \caption{
    \textbf{Visualizations of model behaviors}. We provide five examples from the held-out set with model predictions (red) and ground-truths (green). For each example, we show five frames and a text describing the model behavior below. The translation visualizations (curves) are obtained by projecting the values to the ground and then to the image plane. The rotations (squares) are visualized by projecting the viewing frustums to the image plane. We show the full future sequences for translations while only the next step for rotations for clarity. 
    }
    \label{fig:qualitative}
\end{figure*}

\parahead{Diverse model behaviors}
We evaluate if our trained model demonstrates desirable behaviors through visual inspection. Specifically, we are interested in seeing (1) if our model predicts trajectories that are free from collision with static and dynamic obstacles, (2) if our model learns human-like information-gathering behaviors, and (3) where our model fails. For this purpose, we visualize our model's predictions and ground-truths in 2D, and overlay them over the image observations. We show such visualizations for a few samples in~\cref{fig:qualitative} and more on the project webpage. It can be seen that our model forecasts collision-free paths both around static and dynamic obstacles. Our model also learns the information-gathering behavior that humans demonstrate in the training data, such that it predicts head rotations that check potentially useful information (\eg road conditions) for navigation. We also observe other interesting behaviors from the model. In the first and fifth examples, the model learns to wait when there is no easily traversable path available. In the third example, the model adapts its predictions based on new visual cues it observes (the predicted path shifts from the center to the right after the person appears).

\begin{figure}[t]
    \centering
    \includegraphics[width=\linewidth]{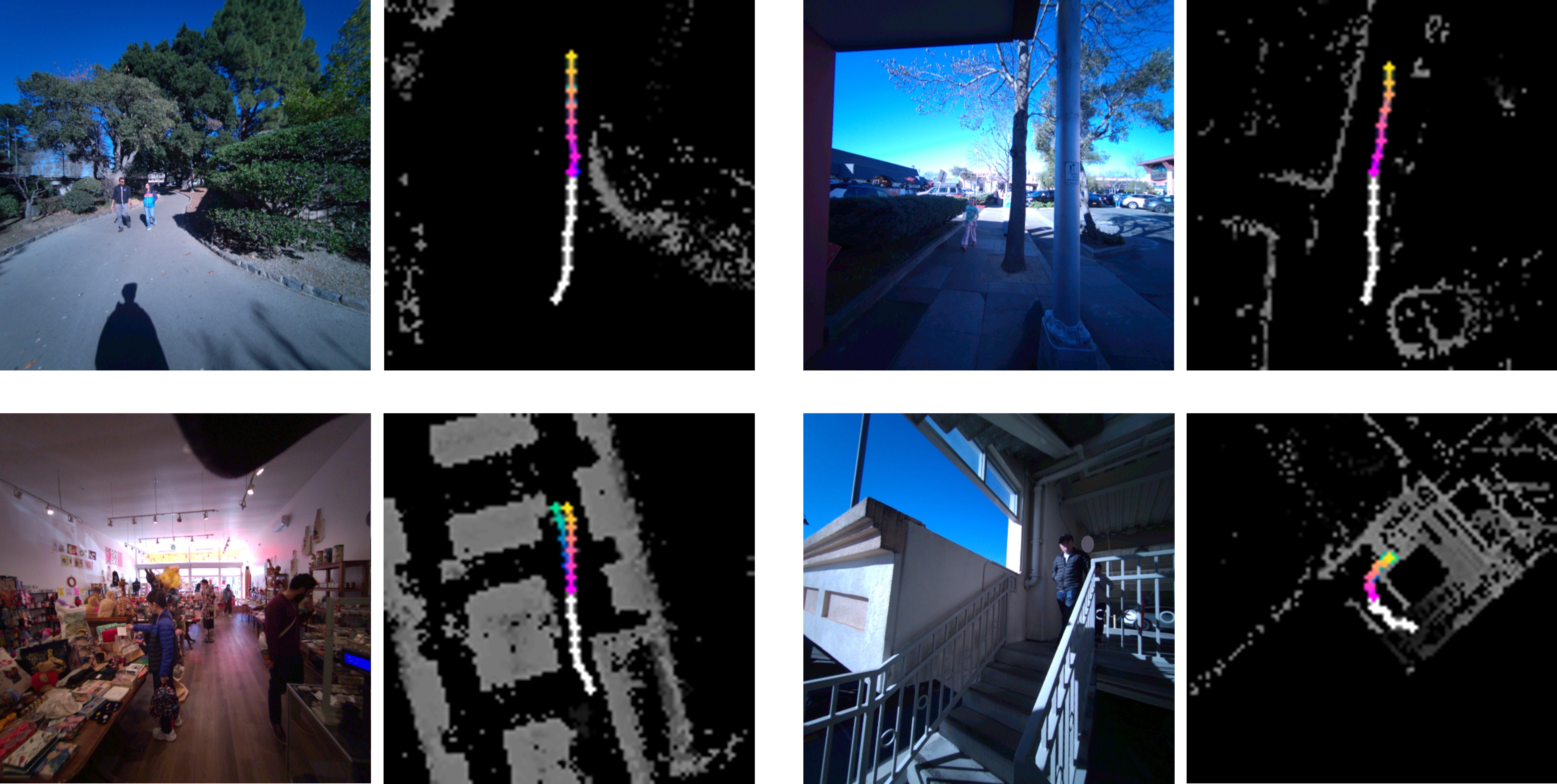}
    \caption{
    \textbf{BEV visualizations}. We show four examples, each with a sampled RGB frame on the left and the BEV visualization of trajectories on the right. The white curve denotes the past, blue-green denotes the ground-truth future, and pink-orange represents the predicted future. Color coding depicts the order of time progression. The trajectories are overlaid on a BEV representation of the scene point cloud for visualization. Note that only the translation components of the trajectories are shown here.
    }
    \vspace{-2mm}
    \label{fig:bev}
\end{figure}

\parahead{BEV visualizations}
We additionally show translation visualizations from the BEV in~\cref{fig:bev}. We overlay the trajectories on top of a BEV representation of the static environment, which is obtained by converting the scene point cloud to an occupancy grid and then to a height map. The height map stores for each pixel, the maximum height of all occupied grids along the up-axis. As seen again from these visualizations, the predicted trajectories from our model satisfy environmental constraints in diverse scenarios.

\subsection{Failure Cases and Limitations}
\label{subsec:failure_cases}

\begin{figure}[t]
    \centering
    \includegraphics[width=\linewidth]{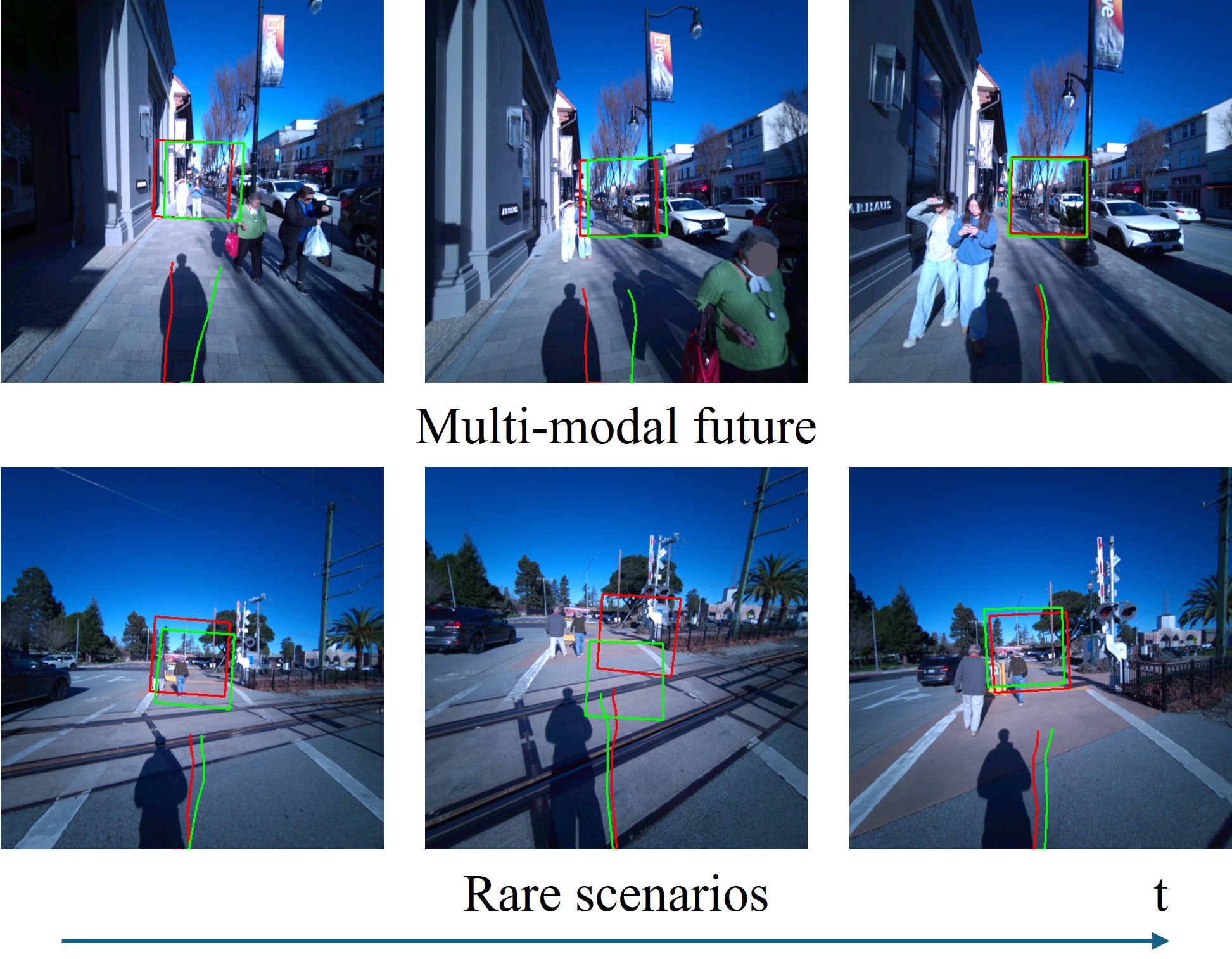}
    \caption{
    \textbf{Failure cases}. We show two examples from the held-out set with a description of the failure below each. 
    }
    \vspace{-2mm}
    \label{fig:failure}
\end{figure}

We identify failure cases of our model and provide the visualizations in~\cref{fig:failure}. A major limitation of our model is its lack of generative modeling capabilities and hence may struggle when a multi-modal future is possible. In the first example of~\cref{fig:failure}, it is possible to go either left or right to avoid collisions with the incoming pedestrians, in which case our model shall regress to the mean of these multiple possibilities. Only when the human subject in this case clearly walks to their right side, our model is able to regress to a plausible future. A next step is thus to leverage generative models to learn such multimodal distributions, such as diffusion models~\cite{DDPM2020,DDIM2020}.
In the second case, the human subject looks down to check the position of the rail in the middle time step to avoid tripping. However, our model does not make such predictions because rails have never appeared in our training set. Expanding our training set to include more diverse scenarios would be a promising solution to this problem.

%% file: content/main_paper/text/6_conclusion.tex
\section{Conclusion}
In this paper, we make steps towards a real-world deployable humanoid navigation policy by making a number of contributions. First, we introduce a novel task of predicting the future 6D head pose trajectory from the past posed egocentric video, under the presence of both static and dynamic obstacles. This task formulation allows the model to learn to not only plan collision-free paths but also learn human-like information-gathering behaviors. Second, we propose a model that leverages pre-trained DINO feature encoders and a parameter-free unprojection strategy to effectively solve this task. Next, we design a data collection pipeline that uses only a pair of Project Aria glasses as the data capture device. This pipeline is easily scalable and allows us to collect a 4-hour real-world navigation dataset with ease. Our dataset spans 18 places with diverse and dense traffic, providing the community with a valuable resource. Through extensive experiments, we demonstrate that our model learns diverse behaviors that are useful for real-world navigation tasks, and surpasses baselines across all metrics. Finally, we discuss the failure cases and limitations of our model as well as directions for future work.

%% file: content/main_paper/text/7_acknowledgment.tex
\subsection*{Acknowledgments}
This work was supported by ARL under grant number W911NF-21-2-0104, a Vannevar Bush Faculty Fellowship, HAI, and Meta Reality Labs.